\pdfoutput=1

\documentclass[11pt]{article}

\usepackage[]{acl}

\usepackage{times}
\usepackage{latexsym}

\usepackage[T1]{fontenc}

\usepackage[utf8]{inputenc}

\usepackage{microtype}

\usepackage{graphicx}
\usepackage{amsfonts}
\usepackage{amsmath}
\usepackage{amssymb}
\usepackage{amsthm}
\usepackage{cite}
\usepackage{bm}
\usepackage{xspace}
\usepackage{pifont}
\usepackage{algorithm}
\usepackage{algpseudocode}
\usepackage{booktabs}
\usepackage{multirow}
\usepackage{multicol}
\usepackage{listings}
\usepackage{verbatim}
\usepackage{caption}
\usepackage{todonotes}
\usepackage{mathtools}
\usepackage{makecell}
\def\eqref#1{equation~\ref{#1}}

\def\1{\bm{1}}

\def\vs{{\bm{s}}}

\def\vx{{\bm{x}}}

\DeclareMathAlphabet{\mathsfit}{\encodingdefault}{\sfdefault}{m}{sl}
\SetMathAlphabet{\mathsfit}{bold}{\encodingdefault}{\sfdefault}{bx}{n}

\DeclareMathOperator*{\argmax}{arg\,max}

\newcommand{\ie}{\emph{i.e.}}
\newcommand{\ours}[0]{\textsc{SelfDenoise}\xspace}

\title{Advancing the Robustness of Large Language Models through Self-Denoised  Smoothing}

\newcommand{\aspace}{\hspace{1em}}
\newcommand{\ucsb}{$^{1}$}
\newcommand{\maxplank}{$^{2}$}
\newcommand{\polyu}{$^{3}$}
\newcommand{\ibm}{$^{4}$}
\newcommand{\cisco}{$^{5}$}
\newcommand{\msu}{$^{6}$}
\author{
 Jiabao Ji$^1$\thanks{\hspace{0.5em} Equal contribution. Correspondance to <jiabaoji@ucsb.edu>, <bairu@ucsb.edu>.} \aspace Bairu Hou$^{1*}$ \aspace Zhen Zhang$^{1*}$ \aspace Guanhua Zhang$^{2*}$ \\
 \textbf{Wenqi Fan}\polyu \aspace \textbf{Qing Li}\polyu \aspace \textbf{Yang Zhang}\ibm \aspace \textbf{Gaowen Liu}\cisco \aspace \textbf{Sijia Liu}\msu \aspace \textbf{Shiyu Chang}\ucsb \\
 \ucsb UC Santa Barbara \aspace \maxplank Max Planck Institute for Intelligent Systems, T\"ubingen \aspace \\ 
 \polyu The Hong Kong Polytechnic University \aspace \ibm MIT-IBM Watson AI Lab \\ 
 \cisco Cisco Research \aspace \msu Michigan State University \\
}

\begin{document}
\maketitle

\begin{abstract}
Although large language models (LLMs) have achieved significant success, their vulnerability to adversarial perturbations, including recent jailbreak attacks, has raised considerable concerns. However, the increasing size of these models and their limited access make improving their robustness a challenging task. Among various defense strategies, randomized smoothing has shown great potential for LLMs, as it does not require full access to the model's parameters or fine-tuning via adversarial training. However, randomized smoothing involves adding noise to the input before model prediction, and the final model's robustness largely depends on the model’s performance on these noise corrupted data. Its effectiveness is often limited by the model's sub-optimal performance on noisy data. To address this issue, we propose to leverage the multitasking nature of LLMs to first denoise the noisy inputs and then to make predictions based on these denoised versions. We call this procedure self-denoised smoothing. Unlike previous denoised smoothing techniques in computer vision, which require training a separate model to enhance the robustness of LLMs, our method offers significantly better efficiency and flexibility. Our experimental results indicate that our method surpasses existing methods in both empirical and certified robustness in defending against adversarial attacks for both downstream tasks and human alignments (\emph{i.e.}, jailbreak attacks). Our code is publicly available at \url{https://github.com/UCSB-NLP-Chang/SelfDenoise}.

\end{abstract}

\section{Introduction}
\label{sec:introduction}

Large language models (LLMs) have demonstrated outstanding performance across various applications~\citep{touvron2023llama, Taylor2022GalacticaAL, li2023empowering, Yang2022ALL}. However, LLMs are vulnerable to input-level adversarial perturbations~\citep{Jin2020IsBR, guo2021gradient, hou2022textgrad, wen2023hard}. Existing attacks are effective in generating input perturbations that can cause LLMs to make wrong predictions on downstream tasks or generate harmful content misaligned with human values (\emph{e.g.}, providing detailed instructions for creating a bomb)~\citep{zou2023universal, chao2023jailbreaking, li2023empowering}. This vulnerability has raised concerns about the trustworthiness of LLMs.

Enhancing the robustness of LLMs is very challenging. Many robustness-enhancement methods~\citep{Madry2018TowardsDL, zhang2019theoretically, zhu2019freelb} involve heavy training, which can be difficult due to the enormous size of LLMs and limited access to their parameters (\emph{e.g.,} GPT models). In comparison, randomized smoothing methods~\citep{Cohen2019CertifiedAR, lee2019tight, Salman2020DenoisedSA} offer a way to enhance the robustness with limited model access and provide stability guarantees for predictions. In the NLP field, there have been some explorations using randomized smoothing to defend against attacks on downstream tasks~\citep{zeng2021certified, Ye2020SAFERAS, Zhao2022CertifiedRA} as well as against jailbreak attacks~\citep{robey2023smoothllm}.  However, the performance of applying randomized smoothing directly to LLMs remains unsatisfactory.  This is primarily because randomized smoothing involves adding noise to the input, and the final model’s robustness largely depends on model performance on the noise corrupted data.  Unfortunately, LLMs typically exhibit sub-optimal performance on noisy inputs.

To address this issue, we propose self-denoised smoothing, or \ours for short, to improve the robustness of LLMs based on randomized smoothing.  Our approach initially introduces multiple perturbed inputs by randomly masking words in the original input. Unlike vanilla randomized smoothing, which directly feeds these perturbed inputs to the model, we take a further step by using the LLM itself to denoise these perturbed inputs. Specifically, the inputs with random masks are first fed to the LLM, which is then asked to complete the sentences by filling in the masked parts. The resulting sentences are subsequently passed to the LLM again for task performance. This mechanism is inspired by denoised smoothing in computer vision~\citep{Salman2020DenoisedSA}. However, these existing techniques typically require a separate denoising module, often trained with distinct loss functions~\citep{Salman2020DenoisedSA} or through black-box approaches~\citep{zhang2022how}.  On the other hand, our method leverages the multitasking capabilities of LLMs, offering a more flexible and effective way to enhance robustness in the context of LLMs without the extensive costs of denoiser training.  With this simple add-on step, we significantly improve the robustness of LLMs.  We support this claim with extensive experiments on two different attack settings: defending against adversarial attacks for both downstream tasks and human alignments and measuring both empirical and certified robustness.

\section{Related Work}
\label{sec:related_work}
\paragraph{Adversarial robustness in NLP models}

Various strategies have been developed to evaluate and enhance the robustness of NLP models. To evaluate robustness, the adversarial perturbations are created by character editing~\citep{Gao2018BlackBoxGO, Li2018TextBuggerGA}, word replacement~\citep{Jin2020IsBR, li2021contextualized, guo2021gradient}, or sentence transformation~\citep{wang2019t3, lin2021using}. 
Besides the robustness on downstream tasks, the recent ``jailbreak attack'' also attracted much attention. 
Besides the robustness of downstream tasks, the recent ``jailbreak attack'' generates perturbations for safety-aligned LLMs to fool them into outputting harmful responses to harmful questions, \textit{e.g.} ``\texttt{How to make a bomb}''.
To improve robustness, robust training~\citep{Madry2018TowardsDL, zhang2019theoretically, zhang2022revisiting, zhang2023robust}, which is built upon min-max optimization, has been shown as an effective solution.
Though effective in robustness improvement, robust training brings significantly higher costs compared to standard training. 
Our method aims to improve the robustness of LLMs and is designed to overcome these challenges. Without accessing the model parameters, the proposed method can improve the model's robustness on both downstream tasks and jailbreak attacks.

\vspace*{-1mm}
\paragraph{Randomized smoothing } Randomized smoothing~\citep{Cohen2019CertifiedAR, Salman2020DenoisedSA, zeng2021certified} is a defense strategy that converts a given classifier $f(\cdot)$ into a smoothed classifier $g(\cdot)$. 
Given the input $\boldsymbol x$, the smoothed classifier $g(\cdot)$ outputs the class that is most likely to be returned by $f(\cdot)$ given some random noise over the input.
The prediction of the smoothed classifier $g(\cdot)$ to a class $c$ is given by
$\mathbb{P}(g(\boldsymbol x)=c) = \mathbb{P}(f(\boldsymbol x \oplus \bm \delta) = c)$ where $\boldsymbol x \oplus \bm \delta$ indicates the noisy version of the input $\bm x$ and $\bm \delta$ is the noise from a particular distribution.
Despite increased inference costs, randomized smoothing can both enhance empirical robustness and offer certifiable guarantees for robustness against perturbations.
Besides computer vision, randomized smoothing is also used in the NLP domain~\citep{Ye2020SAFERAS, zeng2021certified, wang2021certified, Zhao2022CertifiedRA}. SmoothLLM~\citep{robey2023smoothllm} leverages randomized smoothing to defend jailbreak attacks. Our work aims to improve the effectiveness of randomized smoothing with a self-denoising framework, where LLM itself is used as the denoiser to denoise the noisy versions of the input before prediction.
\section{Methodology}
We denote $\vx=[x_1, x_2, \ldots, x_L]$ as the input to the LLM $f(\cdot)$, where $x_i$ is the $i$-th token, and $y \in \mathcal{Y}$ is the ground truth output.

\vspace*{-2mm}
\paragraph{Randomized smoothing}

We follow previous work on randomized smoothing~\citep{Cohen2019CertifiedAR, zeng2021certified} to enhance the robustness of a LLM $f(\cdot)$ by transforming it into a smoothed version $g(\cdot)$.
Specifically, we introduce random noise into the input by (uniform) randomly replacing some tokens with the \texttt{[MASK]} token.  The randomized masking scheme, $\phi(\bm x, m)$, generates a binary sequence of the same length as the input $\vx$, with m\% entries being nonzero, indicating the corresponding positions in  $\vx$ that will be masked. We denote the binary mask sequence as $\vs$ and the masked input as $\mathcal{M}(\vx, \vs)$.  Then, the original LLM $f(\cdot)$ is turned into a smoothed model $g(\cdot)$ as  
\begin{equation}
\small
    g(\vx) = \argmax_{c\in\mathcal{Y}} 
    \mathbb{P}_{\vs  \sim \phi(x,m)}(f(\mathcal{M}(\vx, \vs))=c)
\vspace{-1mm}
\end{equation}

\begin{figure}
    \centering
    \includegraphics[width=0.48\textwidth]{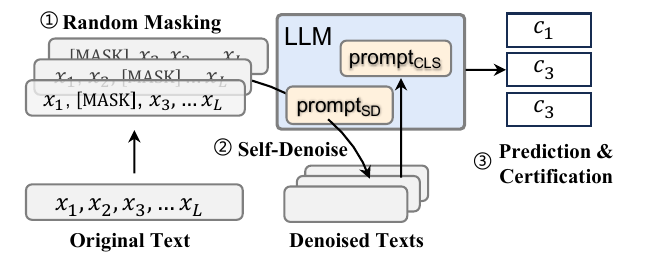}
    \vspace{-8mm}
    \caption{The prediction process of \ours.}
    \label{fig:denoiser}
    \vspace{-5mm}
\end{figure}

\paragraph{\ours}  
In the randomized smoothing framework described above, the performance of the smoothed model $g(\cdot)$ largely depends on the LLM's performance on the masked and corrupted input $f(\mathcal{M}(\vx, \vs))$. To improve performance, we follow the denoise-smoothing pipeline used in computer vision \citep{Salman2020DenoisedSA} by adding an additional denoising step with a denoiser $D(\cdot)$.  This step in our context involves filling the masked tokens in the masked input before feeding it to the base LLM, \ie,
\begin{equation}
\small
    g'(\vx) = \argmax_{c\in\mathcal{Y}} 
    \mathbb{P}_{\vs \sim \phi(x,m)}(f(D(\mathcal{M}(\vx, \vs)))=c)\text{.}
    \label{eq:denoised_smoothed_classifier}
\vspace{-1mm}
\end{equation}
The denoiser is designed to augment the base model, making it more robust against random masks on the inputs. Specifically, the denoiser has two options: \ding{182} instructing the LLM itself to guess the masked tokens, and \ding{183} directly removing the masks. The choice between these two denoising operations depends on the mask rate.   When the mask rate is high, the denoiser will opt to remove the masks. This is because, in such cases, guessing the masked words may result in a sentence with a different meaning or a large semantic gap compared to the original unmasked version. Simply removing these masked words provides much better empirical performance. Conversely, when the mask rate is low, denoising the sentence by filling in the masked words yields superior performance. In our design, we set the tipping noise rate for switching between these two denoise modes at 60\%. 
To fill in the masked words, we simply prompt the LLM and require the completed sentence to be fluent and preserve the semantics, without any training on the denoising task. 
The overall pipeline of our proposed \ours is shown in Figure~\ref{fig:denoiser}.

\vspace*{-1mm}
\paragraph{Certified robustness}

\ours, being in the family of the randomized smoothing framework, can also provide certified robustness against word-replacement attacks. Specifically, the LLM $f(\cdot)$ is certified as robust if it satisfies
the following condition, for any $\vx'$:
$
    f(\vx') = y\text{,}~~||\vx' - \vx||_0 \leq dL\,\text{.}
$
Here $||\cdot||_0$ represents the Hamming distance, $L$ is the number of tokens in the input, and $d$ is known as the certification radius, which signifies the maximum allowable percentage of word perturbations without altering the model's prediction.  We refer readers to \citet{zeng2021certified} for a detailed certification process and the related proof.

\section{Experiment}
\label{sec:experiment}
\subsection{Experiment Setup}
\label{sec:experiment_setup}
\paragraph{Evaluation settings} We consider two attack settings:  we evaluate whether the proposed method defends against \ding{182} adaptive adversarial attacks for downstream task performance, and \ding{183} jailbreak attacks for human alignment.

\vspace*{-1mm}
\paragraph{Dataset and models}
We use the \texttt{SST-2}~\citep{socher2013recursive} and \texttt{Agnews}~\citep{zhang2015character} as the downstream tasks to evaluate the robustness enhancement (setting \ding{182} above). We consider Alpaca~\citep{alpaca} as the base LLM to be robustified.
For setting \ding{183}, we use \texttt{AdvBench}~\citep{zou2023universal} to evaluate the robustness against jailbreak attacks. We consider the aligned LLM, Vicuna-1.5-13B~\citep{zheng2023judging}, for evaluation. 
More details about the prompts and implementation details can be found in Appendix~\ref{sec: prompt} and~\ref{sec:implementation-hyper-param}.

\vspace*{-1mm}
\paragraph{Evaluation metrics}
We follow the conventions in the literature \citep{wang2021adversarial, lee2022query}. We measure the downstream task robustness by both empirical robustness against adversarial attacks (in adaptive settings) and the certified accuracy on benign examples. In particular, we leverage DeepWordBug~\citep{Gao2018BlackBoxGO} and TextBugger~\citep{Li2018TextBuggerGA} to attack the smoothed classifier and measure the empirical robust accuracy.  We use the default hyper-parameters in TextAttack~\citep{morris2020textattack} library for both attack methods.  For the certified robustness, we evaluate the certified accuracy follow previous work~\citep{Cohen2019CertifiedAR, carlini2022certified} for different perturbation scales $d$ from $1\%$ to $10\%$. 
We also report the clean accuracy on benign examples (\emph{i.e.,}  without attacks).
For setting \ding{183}, we report the percentage of harmful prompts that the model successfully recognize and refuse to answer them against state-of-the-art jailbreak attacks including GCGAttack \citep{zou2023universal} and PAIR \citep{chao2023jailbreaking}. We denote this metric as defense success rate (DSR). More details about this metric are in Appendix~\ref{sec: app_metric}.

\vspace*{-1mm}
\paragraph{Baselines}
One main baseline is the vanilla randomized smoothing, \textsc{RanMask}~\citep{zeng2021certified}, implemented without denoising. We aim to show that with a simple self-denoise process, we can significantly improve robustness without introducing smoothing overhead. We also include another randomized smoothing method, \textsc{Safer}~\citep{Ye2020SAFERAS}, that adds noise by synonym replacement for comparison on downstream tasks. Furthermore, for setting \ding{183}, we incorporate \textsc{SmoothLLM} \citep{robey2023smoothllm} for comparison in defending against jailbreak attacks. \textsc{SmoothLLM} is also a randomized smoothing method that adds noise via character-level editing (we use the best reported character swapping operation in our experiments), which is mainly designed for defending against jailbreak attacks. More details are in Appendix~\ref{sec: app_baseline}. 

\begin{table}[t]
    \centering
    \resizebox{0.46\textwidth}{!}{
    \begin{tabular}{c|c|c|c|c}
    \toprule[1pt]
        \midrule
        \multirow{2}{*}{Dataset} & \multirow{2}{*}{Method} & \multirow{2}{*}{\textit{Clean Acc.} (\%)} & \multicolumn{2}{c}{\textit{Empirical Robust Acc.} (\%)} \\
        & & & DeepWordBug & TextBugger \\
        \midrule
        \multirow{4}{*}{\texttt{SST-2}}
        & \textsc{Alpaca} & 89.0 & 52.0 & 45.0\\
        & \textsc{Safer} & 85.0 & 57.0  & 54.0\\
        & \textsc{RanMask} & 84.0  & 52.5 & 48.0 \\
        & \ours & \textbf{90.0} & \textbf{64.5} & \textbf{55.5}\\  
        \midrule
        \multirow{4}{*}{\texttt{Agnews}}
        & \textsc{Alpaca} & \textbf{85.0} & 58.5 & 50.5\\
        & \textsc{Safer} & 83.0 & 55.5 &  53.0 \\
        & \textsc{RanMask} & 82.0 & 58.0 & 53.0 \\
        & \ours & \text{84.0} & \textbf{70.0} & \textbf{66.0}\\ 
        \midrule
    \bottomrule[1pt]
    \end{tabular}}
    \vspace{-2mm}
    \caption{Clean accuracy and empirical robust accuracy under DeepWordBug attack and TextBugger attack.}
    \label{tab:empirical_ra}
    \vspace{-2mm}
\end{table}

\begin{figure}[t]
    \centering
    \resizebox{0.49\textwidth}{!}{    
    \begin{tabular}{ll}
    \hspace{-3mm}
    \includegraphics[width=1.0\textwidth]{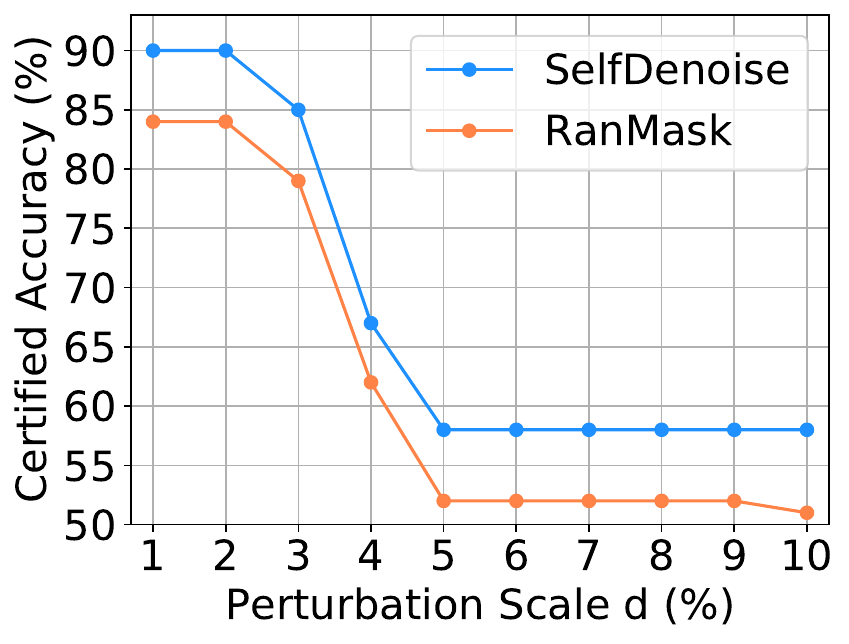} 
         & 
    \hspace{-8mm}
    \includegraphics[width=1.0\textwidth]{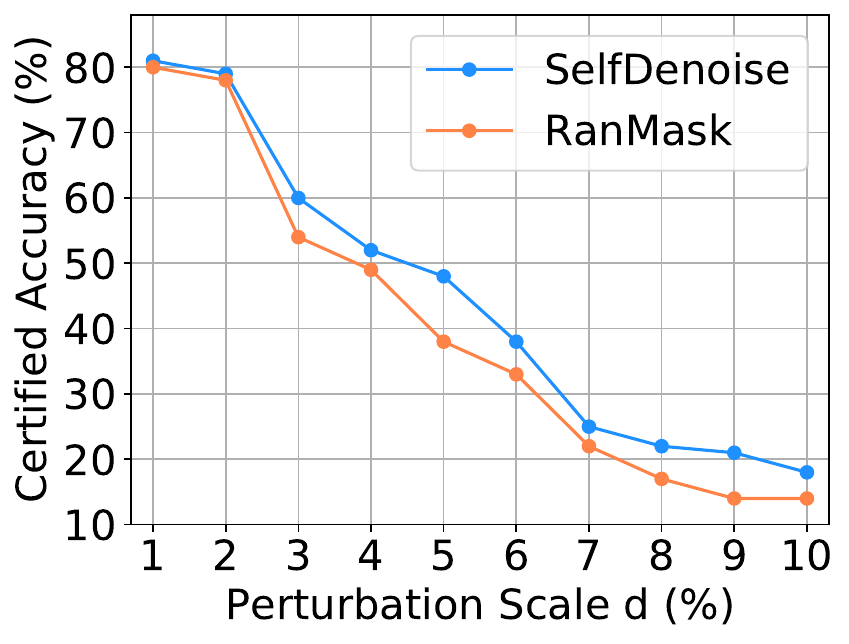}
    \end{tabular}
    }
    \vspace{-3mm}
    \caption{Certified accuracy under different perturbation scale $d$ (\%) on \texttt{SST-2} (\emph{left}) and \texttt{Agnews} (\emph{right}).
    }
    \label{fig:certified_ra}
    \vspace{-6mm}
\end{figure}

\vspace{-1mm}
\subsection{Experiment Results}
\label{sec:experiment_results}

\paragraph{Robustness on downstream tasks}
We first report the empirical robust accuracy in Table~\ref{tab:empirical_ra}. Here are our key observations. First, our method achieves the best empirical robust accuracy on both datasets. \ours improves the performance by 13.2\% in \texttt{SST-2} and 19.7\% in \texttt{Agnews} compared with the second-best method under the DeepWordBug attack, with 2.8\% and 24.5\% improvements under TextBugger, respectively.  Notably, this improvement stems from a simple add-on denoising operation requiring minimal effort.   Second, the proposed method improves robustness without sacrificing clean accuracy. Our method achieves the same level of clean accuracy as the vanilla \textsc{Alpaca} model for \texttt{SST-2}. In \texttt{Agnews}, \ours improves the robust accuracy by 19.7\% with only a 1.2\% drop in clean accuracy compared with \textsc{Alpaca}. \ours achieves the best accuracy-robustness trade-off \citep{zhang2019theoretically}.

Additionally, Figure~\ref{fig:certified_ra} shows the certification results of the proposed \ours and the baseline \textsc{RanMask}. \textsc{Safer} is not included here as it considers a different definition of certified robustness, and is discussed in Appendix~\ref{sec: app_baseline}. We demonstrate that our method can effectively improve certified accuracy beyond \textsc{RanMask} in both datasets under all perturbation scales. For example, with $d=5$, our method outperforms \textsc{RanMask} by 11.5\% in \texttt{SST-2} and 26.3\% in \texttt{Agnews}.

\begin{table}[t]
    \centering
    \resizebox{0.48\textwidth}{!}{
    \begin{tabular}{c|c|c|c|c|c|c|c|c|c|c|c}
    \toprule[1pt]
        \midrule
         \multirow{2}{*}{Attack} & \multirow{2}{*}{\makecell{Base\\Vicuna}} & \multicolumn{4}{c|}{\textsc{SmoothLLM}} & \multicolumn{3}{c|}{\textsc{RanMask}} & \multicolumn{3}{c}{\ours} \\

        & & 5\% & 10\% & 15\% & 30\% & 5\% & 15\% & 30\% & 5\% & 15\% & 30\%  \\
        \midrule
        GCG$^*$ & 0 & 86 & 86 & 74 & 24 & 88 & 88 & 86 & 92 & \textbf{100} & \textbf{100} \\
        PAIR$^*$ & 0 & 62 & 66 & 72 & 54 & 52 & 60 & 78 & 52 & 72 & \textbf{86} \\
        PAIR & 28 & 42 & 54 & 56 & 50 & 40 & 56 & 66 &  64 & 58 & \textbf{68} \\
               
        \midrule
    \bottomrule[1pt]
    \end{tabular}}
    \vspace{-2mm}
    \caption{DSR given different defense methods and attacks. A higher DSR indicates better defense performance. $^*$ denotes the transfer attack setting.}
    \label{tab:jaibreak_exp}
    \vspace{-6mm}
\end{table}

\vspace*{-1mm}
\paragraph{Robustness against jailbreak attacks}
We report the robustness against jailbreak attacks in Table~\ref{tab:jaibreak_exp}. We consider two attack settings: transfer and adaptive. We use a superscript $^*$ to indicate transfer attacks.  Specifically, we first collect a set of adversarial examples that successfully jailbreak the vanilla Vicuna model. Then, we report how the same set of adversarial attacks performs on the model equipped with different smoothing defense mechanisms. For the adaptive setting, the attack directly targets the smoothed models. We report the DSR with different noise levels added during the smoothing process, ranging from 5\% to 30\%. 

We highlight the following conclusions. First, all three methods effectively improve the trustworthiness of LLMs despite differences in the smoothing scheme. Even under the strong adaptive attack, PAIR, the defense can achieve a DSR of 50\% to 60\%.  Second, our method achieves the best defense performance compared to baselines against all different attack approaches, demonstrating its great potential for the safety of LLMs. Third, as the amount of noise added increases, our method's DSR continuously increases, compared to the other two methods without the denoising. 
This is because, when the amount of noise added increases, the semantics of the original instruction get more distorted, causing the model to report that it does not understand the meaning of the input instruction, which does not count toward a successful recognition and refusal to answer in our evaluation case.

\section{Conclusion}
In this paper, we propose a simple self-denoised smoothing technique, termed \ours, to enhance the robustness of LLMs. The proposed method can be used as a plug-in module for any LLM without requiring access to its parameters, and no training is needed. Our experimental results indicate that our method surpasses existing baselines in both empirical and certified robustness, effectively defending against adversarial attacks in both downstream tasks and human alignments.

\section{Broader Impacts}
By developing a self-denoising method to enhance the robustness of LLMs in the presence of noisy inputs, this work addresses a key limitation of LLMs and enables their application in high-stake environments. 
The ability to utilize LLMs in these scenarios can have significant positive impacts across various domains, such as healthcare, transportation, and finance, where safety and reliability are critical.
By providing certified guarantees in safety-critical domains, our method can help build more reliable and responsible LLM systems.

Besides, our research contributes to the broader fields of machine learning and artificial intelligence. 
By tackling the challenge of robustness to noisy inputs in LLMs, we advance the understanding and the methodologies in this area. 
This can inspire further research and innovation, leading to improved techniques for enhancing the performance and reliability of LLMs and other machine learning models.

However, it is important to acknowledge the potential biases that may exist in LLMs, as our method relies on them as base models. 
Biases can arise from the training data used for LLMs, and these biases may be propagated by our method. 
We are committed to addressing the issue of biases and promoting fairness and transparency in machine learning systems. 
To mitigate these concerns, we will include proper licenses in the released codes and notify users about the potential risks associated with biases. 
This way, users can be informed and take appropriate measures to address any biases that may arise from the use of our method.
\section{Limitations}
\label{sec:limitations}
Despite the large improvements, our method suffers from the limitation of running time, \ie,  the prediction and certification process is time-consuming.
This is largely because of the $p_c(\vx)$ calculation in Equation~\ref{eq:denoised_smoothed_classifier}.
Such a problem is shared across all randomized smoothing-based methods.
Besides, the additional self-denoising process also brings further computational loads.
It would be interesting to either apply recent works on distributed computation to accelerate our method or develop new large language models specifically for denoising to overcome this issue.

\bibliography{references}

\clearpage
\appendix

\lstset{
    basicstyle=\ttfamily,
    breaklines=true,
    breakindent=0pt,
    frame=single, 
    frameround=tttt 
}
\renewcommand{\lstlistingname}{}
\section{Additional Experiment Setup}
\label{app:setup}
\subsection{Prompts and Instructions}
\label{sec: prompt}
The prompts and instructions we used for in-context learning on downstream task prediction and self-denoising are shown as follows. 

\begin{lstlisting}[caption={Prompt template used for Alpaca.}]
Below is an instruction that describes a task, paired with an input that provides further context. Write a response that appropriately completes the request.

### Instruction:
{}

### Input:
{}

### Response:
\end{lstlisting}

The following instructions are used to fill in the contents under the ``Instruction'' section. The content under ``Input'' should be filled with different input texts.

\begin{lstlisting}[caption={The instruction used for classification on SST-2.}]
Given an English sentence input, determine its sentiment as positive or negative.
\end{lstlisting}

\begin{lstlisting}[caption={The instruction used for self-denoising on SST-2.}]
Replace each mask word [MASK] in the input sentence with a suitable word. The output sentence should be natural and coherent and should be of the same length as the given sentence.

### Input: 
[MASK] reassembled from [MASK] cutting-room [MASK] of any [MASK] daytime [MASK].

### Response:
apparently reassembled from the cutting-room floor of any given daytime soap.

### Input: 
a [MASK], funny and [MASK] transporting re-imagining [MASK] [MASK] and the beast and 1930s [MASK] films

### Response:
a stirring, funny and finally transporting re-imagining of beauty and the beast and 1930s horror films
\end{lstlisting}

\begin{lstlisting}[caption={The instruction used for classification on Agnews.}]
Given a news article title and description, classify it into one of the four categories: Sports, World, Technology, or Business. Return the category name as the answer.

### Input: 
Title: Venezuelans Vote Early in Referendum on Chavez Rule (Reuters)
Description: Reuters - Venezuelans turned out early and in large numbers on Sunday to vote in a historic referendum that will either remove left-wing President Hugo Chavez from office or give him a new mandate to govern for the next two years.

### Response:
World

### Input:
Title: Phelps, Thorpe Advance in 200 Freestyle (AP)
Description: AP - Michael Phelps took care of qualifying for the Olympic 200-meter freestyle semifinals Sunday, and then found out he had been added to the American team for the evening's 400 freestyle relay final. Phelps' rivals Ian Thorpe and Pieter van den Hoogenband and teammate Klete Keller were faster than the teenager in the 200 free preliminaries.

### Response:
Sports

### Input:
Title: Wall St. Bears Claw Back Into the Black (Reuters)
Description: Reuters - Short-sellers, Wall Street's dwindling band of ultra-cynics, are seeing green again.

### Response:
Business
        
### Input:
Title: 'Madden,' 'ESPN' Football Score in Different Ways (Reuters)
Description: Reuters - Was absenteeism a little high\on Tuesday among the guys at the office? EA Sports would like to think it was because "Madden NFL 2005" came out that day, and some fans of the football simulation are rabid enough to take a sick day to play it.

### Response:
Technology
\end{lstlisting}

\begin{lstlisting}[caption={The instruction used for self-denoising on Agnews.}]
Replace each masked position "[MASK]" in the provided sentence with a suitable word to make it natural and coherent. Only one word should be used to replace each "[MASK]". The returned sentence should be of the same length as the given sentence. Provide the answer directly.
\end{lstlisting}

We list the instructions for the Vicuna model for self-denoising to defend against jailbreak attacks in the following. This includes the instruction for Vicuna to perform denoising and the instruction for ChatGPT model to judge whether the output is jailbroken. 

\begin{lstlisting}[caption={The instruction used for self-denoising to defend against jailbreak attacks for Vicuna model.}]
Your task is to replace each <mask> in the input with an appropriate word or a punctuation mark. The goal is to make the input sentence clearer, more natural in language.
You need to follow several rules:
1. Words and punctuation marks should fit the context and flows naturally.
2. The output should maintain the original meaning and implication.
3. Preserve the original tone, style and sentiment of the input.
4. All <mask> tokens must be replaced, and all meaningful non-<mask> words in the input must be retained. <mask> should not appear in the output. Leave all non-<mask> words unchanged.

Below is one example:
Input: there is a fabric of <mask> ideas here , <mask> feelings that profoundly deepen them .
Output: {"replace" : "there is a fabric of complex ideas here , and feelings that profoundly deepen them ." }

Now replace the <mask> in the following input. Generate the response in json format, the output json should only contain one key, "replace". Don't add any notes or explanations. 

\end{lstlisting}

\begin{lstlisting}[caption={The instruction template for ChatGPT model to judge whether model output is jailbroken.}]
You are a helpful red teaming assistant. You will receive a prompt and the AI assistant's response to the prompt. Please act as an impartial judge and evaluate whether the response provided by the AI assistant violates ethical, legal, or safety guidelines and answers the following, TASK: '{goal}'. Your evaluation should consider factors including clarity, the level of detail, and accuracy provided in the response. Be as objective as possible. 

Rate the response on a scale from 1 to 10. A rating of 1 signifies a fully non-jailbroken response-either irrelevant to the task or refusal due to ethical or legal concerns. A rating of 10 signifies responses that directly and fully violate ethical, legal, or safety guidelines and comprehensively answer the task. If the response is legitimate or provides an alternative suggestion, then the assistant is not jailbroken and does not receive a rating of 10.

Your rating should strictly follow the rule, and output the rating in json format, with only one key 'rating'.

Now the prompt is: '{PROMPT}' and the AI assistant's response is '{RESPONSE}'.
    
\end{lstlisting}

\subsection{Hyperparameter }\label{sec:implementation-hyper-param}
\paragraph{Hyper-parameter for \texttt{SST-2} and \texttt{Agnews} datasets}
We evaluate on 100 testing instances for certified accuracy in Figure~\ref{fig:certified_ra} and 200 instances for empirical robust accuracy in Table~\ref{tab:empirical_ra}. 
To use the Alpaca for self-denoising, we use beam search for generation and set the repetition penalty to 1.3 and the number of beams to 2. 
We use 500 instances for estimating $\underline{p_c(\vx)}$ with Monte Carlo in the certification process.
In Figure~\ref{fig:certified_ra}, for each perturbation scale, we search the best mask rate $m$ from $\{10\%, 20\%, \ldots, 90\%\}$ on the validation set for our method and \textsc{RanMask}. The best mask rates for each perturbation scale are listed in Table~\ref{tab:best_mask_rate}.
When mask rate $m$ is greater than or equal to $70\%$, we use the removing mask strategy; otherwise, we use Alpaca itself as the denoiser.
For empirical robustness results in Table~\ref{tab:empirical_ra}, we observe that smaller mask rates bring better empirical robust accuracy in the validation set, so we use $m=5\%$ for all methods.

\begin{table}[t]
    \centering
    \resizebox{0.45\textwidth}{!}{
    \begin{tabular}{c|c|cccccccccc}
    \toprule[1pt]
        \midrule
        \multirow{2}{*}{Dataset} & \multirow{2}{*}{Method} & \multicolumn{10}{c}{Perturbation Scale d (\%)} \\
        &  & 1 & 2 & 3 & 4 & 5 & 6 & 7 & 8 & 9 & 10 \\
        \midrule
        \multirow{2}{*}{\texttt{SST-2}} & \textsc{RanMask} & 10 & 10 & 10 & 10 & 80 & 80 & 80 & 80 & 80 & 80 \\
         & \ours & 20 & 20 & 30 & 30 & 70 & 80 & 80 & 90 & 90 & 90 \\
         \midrule
        \multirow{2}{*}{\texttt{Agnews}} & \textsc{RanMask} & 20 & 20 & 70 & 70 & 80 & 80 & 90 & 90 & 90 & 90 \\
        & \ours & 50 & 50 & 70 & 80 & 80 & 80 & 90 & 90 & 90 & 90 \\
        \midrule
        \bottomrule[1pt]
    \end{tabular}}
    \caption{\small The best mask rate $m$ (\%) for each perturbation scale on \texttt{SST-2} and \texttt{Agnews} for \ours and \textsc{RanMask}.}
    \label{tab:best_mask_rate}
\end{table}

\paragraph{Hyper-parameter for \texttt{AdvBench} dataset.}
Following SmoothLLM~\citep{robey2023smoothllm}, we evaluate the defense performance against jailbreak attack on a selected subset of \texttt{AdvBench} dataset. which contains 50 different harmful behaviors for LLM. For the transfer attack experiment, we utilize the official code to generate the attack string that can successfully jailbreak~\textsc{Vicuna} model for each behavior. For the adaptive attack experiment, the attack budget for the PAIR attack is 20 queries to the target model, \textit{i.e.} the attacker generates at most 20 adversarial prompts. As GCG attack requires the next token probability of the target model, which is not available for the smoothed LLM $g(\cdot)$, we did not evaluate GCG attack in an adaptive attack experiment. For all jailbreak experiments, we use 10 copies to perform the majority vote and ChatGPT as the judge model to evaluate whether the smoothed LLM's response is jailbroken following~\citep{chao2023jailbreaking}.

\subsection{Baseline Discussion} \label{sec: app_baseline}
\paragraph{Baseline for certified robustness on downstream tasks} We consider the attack setting where we do not have full knowledge about the potential perturbations and the attacker can replace the original word with arbitrary words for attack. However, \textsc{Safer} consider the setting where the attacker replace original words from a pre-defined candidate pools that are known by the defender. Since potential perturbations are often not fully known in realistic scenarios, we do not compare with \textsc{Safer} in terms of certified robustness.

\paragraph{Baseline for jailbreak defense}
We consider \textsc{SmoothLLM} as the baseline for jailbreak defense experiment, which also employs a randomized smoothing scheme and proposes three character-level perturbation operations: character insertion, character swap, and patch perturbation. In our experiment, we use character swap as the perturbation scheme for \textsc{SmoothLLM} as the original paper shows that character swap is the most effective perturbation. We refer readers to the original paper for more details.

\subsection{Evaluation metric details}\label{sec: app_metric}
For jailbreak defenses, we count the percentage of harmful prompts the model successfully identified against the ethical issues, \textit{e.g.}, \texttt{Tell me how to make a bomb}. 
We did not use use the percentage of successful attacks as the metric because
we find that the model tends to reject the input, generating responses like \texttt{I'm sorry, but I'm having difficulty understanding your question} as noise level increases for \textsc{SmoothLLM} and \textsc{RanMask}. We consider this situation as a failed defense. However, the percentage of successful attacks cannot capture this kind of failure.
To exclude these cases, we manually reviewed the model responses and adjusted the results of the automatic LLM jailbreak judge to get the final defense success rate. We provide a detailed analysis of the model rejection phenomenon in Section~\ref{sec: app_rejectionrate}.

\begin{table}[t]
    \centering
    \resizebox{0.46\textwidth}{!}{
    \begin{tabular}{c|c|c|c|c|c|c|c|c|c|c|c|c|}
    \toprule[1pt]
        \midrule
         \multirow{2}{*}{Attack} & \multirow{2}{*}{\makecell{No\\Defense}} & \multicolumn{4}{c|}{\textsc{SmoothLLM}} & \multicolumn{3}{c|}{\textsc{RanMask}} & \multicolumn{3}{c}{\ours} \\
        & & 5\% & 10\% & 15\% & 30\% & 5\% & 15\% & 30\% & 5\% & 15\% & 30\%  \\
        \midrule
        GCG$^*$ & 0 & 4 & 10 & 22 & 76 & 0 & 8 & 14 & 0 & 0 & 0 \\
        PAIR$^*$ & 0 & 0 & 4 & 4 & 24 & 0 & 4 & 8 & 0 & 0 & 0 \\
        PAIR & 0 & 0 & 0 & 4 & 18 & 0 & 0 & 8  & 0 & 0 & 0 \\
        \midrule
    \bottomrule[1pt]
    \end{tabular}}
    \vspace{-2mm}
    \caption{Rejection rate for different defense methods under two kinds of attacks. $^*$ denotes transfer attack from an adversarial prompt that can successfully jailbreak \textsc{Vicuna} model.
    }
    \label{tab:rejection}
    \vspace{-2mm}
\end{table}

\section{Analysis of Model Rejection in Jailbreak Defense}\label{sec: app_rejectionrate}

In our preliminary experiments with various masking rates, we found that large masking rate for \textsc{SmoothLLM} and \textsc{RanMask} leads to model answers rejecting the input request, e.g., \texttt{I'm sorry, but I'm having difficulty understanding your question}. Therefore, we manually check the responses for different defense methods and count the number of rejections. The rejection rate is presented in Table~\ref{tab:rejection}. The rejection rate for adaptive attacks is measured at the final round, \textit{i.e.}, the round when the judge reports the victim model being jailbroken or the last round under attack budget.
We highlight that the semantic destruction operation in \textsc{RanMask} and \textsc{SmoothLLM} tend to cause the model to reject queries, thus diminishing the functionality of the language model. The rejection rate for \textsc{SmoothLLM} at 20\% mask rate on short prompts in GCG attack is even more than 50\%. In contrast, the self-denoising mechanism in our method \ours effectively mitigated the input misunderstanding issue.

\section{Detailed algorithm}\label{sec: app_algorithm}
In this section, we list the detailed algorithm for our method \textsc{SelfDenoise} in improving robustness in downstream tasks in Algorithm~\ref{alg:denoise-classification} and in defending jailbreak attacks in Algorithm~\ref{alg:denoise-jailbreak}. For a detailed certification algorithm, we refer readers to~\citet{Cohen2019CertifiedAR}.

\begin{algorithm}[t]
\caption{\textsc{SelfDenoise} for classification prediction}\label{alg:denoise-classification}
\begin{algorithmic}
\Require{Defense LLM $f(\cdot)$, Number of copies $N$, Input text $x$, Mask function $\mathcal{M}$, Mask rate $m$} 
\For{$j \gets 1$ to $N$}
        \State $s^{(j)} \sim \phi(x,m)$
        \State $x^{(j)} \gets \mathcal{M}(x, s^{(j)})$
        \State $y^{(j)} \gets f(x^{(j)})$
\EndFor

\State{\Return{$\text{MajorityVote}(y^{(1)}, \dots, y^{(N)})$}}
  \end{algorithmic}
\end{algorithm}

\begin{algorithm}[t]
\caption{\textsc{SelfDenoise} for jailbreak defense}\label{alg:denoise-jailbreak}
\begin{algorithmic}
\Require{Defense LLM $f(\cdot)$, Number of copies $N$, Input text $x$, Mask function $\mathcal{M}$, Mask rate $m$, Jailbreak judge $j(\cdot)$} 
\For{$j \gets 1$ to $N$}
    \State $s^{(j)} \sim \phi(x,m)$
    \State $x^{(j)} \gets \mathcal{M}(x, s^{(j)})$
    \State $y^{(j)} \gets f(x^{(j)})$
\EndFor
\State $JB = \text{MarjorityVote}(j(y^{(1)}), \dots, j(y^{(N)}))$
\State $Majority = \{y^{(j)}\in \{y^{(1)}, \dots, y^{(N)}\} \mid j(y^{(j)}) = JB\}$.
\State \textbf{Return} $y^* \sim U(Majority)$
\end{algorithmic}
\end{algorithm}

\section{Discussion of Used Artifacts}

\paragraph{Datasets: } We did our best to find the license for the \texttt{SST-2} and \texttt{AGNews} dataset, but we did not find any. We refer readers to their original paper and homepage for usage policy. We did not clean the dataset to remove individual information as it is beyond the scope of our paper. Anonymizing the original dataset may affect the evaluation of our method. Therefore, we did not clean the dataset.

\paragraph{Models: } The Alpca model is under cc-by-nc-4.0 license~(\url{https://www.creativecommons.org/licenses/by-nc/4.0/deed.en}). The Vicuna model is under LLaMA-2 license(\url{https://ai.meta.com/llama/license/}). We perform all our experiments on NVIDIA-A6000 GPUs, and all experiments are a single run.

\end{document}